\documentclass[letterpaper, 10 pt, conference]{ieeeconf}

\usepackage{color}
\usepackage{amsmath}
\usepackage{amssymb}
\usepackage{algorithmic}
\usepackage{graphicx}
\usepackage{fixltx2e}
\usepackage{booktabs}
\usepackage{subcaption}
\usepackage{siunitx}
\usepackage{xcolor}
\usepackage{hyperref}

\newcommand{\etal}{et al.~}
\graphicspath{ {figures/}}

 \IEEEoverridecommandlockouts
\begin{document}

\title{An Integrated Simulator and Dataset that Combines Grasping and Vision for Deep Learning}

\author{Matthew Veres, Medhat Moussa, and Graham W.~Taylor
\thanks{This work is supported by the Natural Sciences and Engineering 
Research Council of Canada, and the Canada Foundation for Innovation.}
\thanks{
Authors are with the School of Engineering,
        University of Guelph, 50 Stone Road East. Guelph, Ontario, Canada.
        {\tt\small \{mveres,mmoussa,gwtaylor\}@uoguelph.ca}}%
}

\maketitle

\begin{abstract}
Deep learning is an established framework for learning hierarchical data
representations. While compute power is in abundance, one of the main challenges
in applying this framework to robotic grasping has been obtaining the amount of
data needed to learn these representations, and structuring the data to the task
at hand. Among contemporary approaches in the literature, we highlight key
properties that have encouraged the use of deep learning techniques, and in this
paper, detail our experience in developing a simulator for collecting
cylindrical precision grasps of a multi-fingered dexterous robotic hand.
\end{abstract}

\textbf{\textit{Index Terms ----} Grasping; Barrett Hand; Simulator; Vision; Data Collection;}


\IEEEpeerreviewmaketitle%

\section{Introduction}

Grasping and manipulation are important and challenging problems in Robotics.
For grasp synthesis or pre-grasp planning, there are currently two dominant
approaches: analytical and data-driven (i.e.~learning). \textit{Analytic}
approaches typically optimize some measure of force- or
form-closure~\cite{sahbani2012overview} \cite{bohg2014data}, and provide
guarantees on grasp properties such as: disturbance rejection, dexterity,
equilibrium, and stability \cite{suarez2006grasp}. These models often require
full knowledge of the object geometry, surface friction, and other intrinsic
characteristics. Obtaining these measurements in the real world is difficult,
and measurements are often imperfect due to sensor limitations, including noise.
A different approach that has recently gained significant interest is the
\textit{data-driven} or \textit{learning} approach. In this case, the emphasis
is placed on learning from data how to ``best'' grasp an object, which affords
significant flexibility and robustness in uncertain real-world environments.
Many learning algorithms have been proposed
\cite{bohg2014data,moussa1998experimental}, and most recently have included
algorithms within the deep learning framework.

\subsection{The challenges of data-driven approaches}

Obtaining data for learning how to grasp is very difficult. There are many
reasons for this difficulty, including: access to physical resources needed to
run robotic experiments continuously, and the time it takes to collect a large
dataset. The data collection process itself is not standard, and there is no
clear experimental process that accounts for the infinite variability of
manipulators, tasks, and objects. If deep learning is used, this problem is only
magnified as these sets of models and learning algorithms are known to require
significantly larger amounts of data. Nonetheless, there are several initiatives
to collect data from grasping experiments on a large scale. Pinto and Gupta
~\cite{pinto2015supersizing} were able to collect over 700 hours worth of
real-world grasps using a Baxter robot. A similar initiative by Levine
\etal\cite{levine2016learning} has explored data collection through robotic
collaboration --- collecting shared grasping experience across a number of
real-world robots, over a period of two months.

Alternative environments for large-scale data collection also exist.
\textit{Simulators} alleviate a significant amount of real-world issues, and are
invaluable tools that have been accelerating research in the machine learning
community. Recent works leveraging simulated data collection for robotic
grasping include Kappler \etal\cite{kappler2015leveraging}, who collect over
300,000 grasps across 700 meshed object models, and Mahler
\etal\cite{mahlerdex}, who collected a dataset with over 2.5 million
parallel-plate grasps across 10,000 unique 3D object models.

Our long-term objective is to explore learning approaches and representations
that combine object perceptual information with tactile feedback, to enable
grasping under various object characteristics and environmental conditions. This
requires the  simulation of robotic grasps using a variety of different
grippers, different object shapes and characteristics, and many different
sensory systems, each capturing different parts of the grasping process.

There are a number of different robotic simulators that have emerged over the
years, such as OpenRAVE~\cite{diankov_thesis},  ROS/Gazebo
\cite{quigley2009ros}, Graspit!~\cite{miller2004graspit}, and V-REP~\cite{VREP}.
For interested readers, Ivaldi \etal\cite{ivaldi2014tools} carried out a
user-based survey of tools for robotic simulation currently in use, and
\cite{erez2015simulation} provides an interesting comparison of different tools.
In this work, we use V-REP for its capability of rapid prototyping, range of
supported sensors, and flexible choice of dynamics engine.

\subsection{Paper contribution}

In~\cite{veres_2017} we presented an integrated object-action representation
that we call grasp motor image. We demonstrated its capacity for capturing and
generating multimodal, multi-finger grasp configurations on a simulated grasping
dataset. In this paper, we provide more details about the integrated simulation
environment that was used in~\cite{veres_2017}\footnote{Note that there has been
some minor changes between the simulation used in \cite{veres_2017} and the
simulator introduced here, largely with respect to the collected information
(e.g.~image size) and objects used.}. Leveraging the multifaceted nature of
V-REP and the plethora of sensors available, this environment enables grasping
experience and object perceptual properties to be captured together, during the
process of grasping. We provide this simulation and associated code as an
open-source resource to the community, along with a collected dataset that
contains over 50,000 successful grasps, split across 64 classes of objects.
Should anyone wish to develop their own simulation, we outline in the remainder
of this paper some considerations we chose, along with an example of how this
simulation can be run across many compute nodes for collecting data in parallel.

\section{Simulation Overview and Architecture}

We chose to create our simulation with two key ideas in mind: (1) A grasp can be
represented in a generic manner through an object-centric reference frame, and
(2) Grasp candidates can be sampled through the simple application of pre- and
post- multiplication of rotation matrices.

\begin{figure}
	\centering
	\includegraphics[width=.95\linewidth]{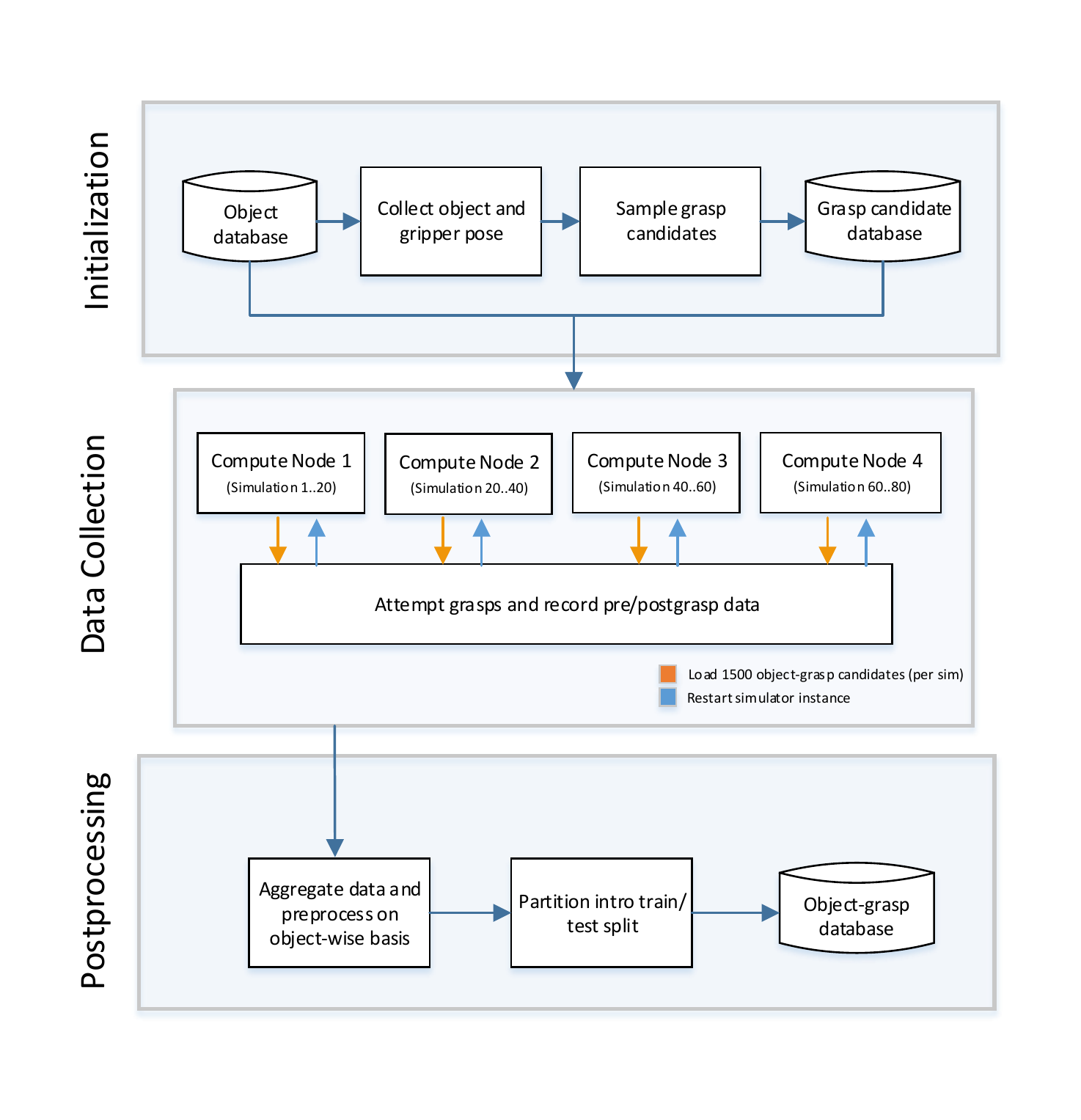}
	\caption{Overview of dataset preparation pipeline}
\label{fig:simulator_pipeline_overview}
\end{figure}

Each simulation consists of three stages: i) pre-processing which includes 
initializing object parameters and generating grasp candidates, ii) executing a 
simulation task and collecting data, and iii) postprocessing the collected data. 
These stages are discussed in depth in Section~\ref{sec:pipeline}, and a general
overview is presented in Figure~\ref{fig:simulator_pipeline_overview}.

\subsection{V-REP simulation environment}

The native programming language of V-REP is Lua, and the most direct approach
for customizing simulations is to write \textit{embedded scripts}. These scripts
are fully-contained within the simulator, are platform independent, and fully
compatible with other V-REP installations~\cite{vrepWritingCode}. It is also
possible to customize through auxiliary methods, such as through add-ons,
plugins, various remote APIs or ROS nodes. We chose to use embedded scripts, as
development was being done between Windows and Linux environments, and for
future work with parallelization allowed us to circumvent additional
communication lag or processing overhead.

One of the features of V-REP is that the entire task can be simulated.  Grasping
is an intermediate operation in an overall robotics task; a simulator that can
simulate the entire task starting from perception would be more realistic.  This
process also includes other factors such as obstacles around the object, as well
as reachability and singularity constraints. V-REP supports integrated
path-planning and obstacle avoidance modules, as well as inverse kinematics and
support for a wide range of manipulators, grippers, and object types.

A variety of sensors (including both tactile and perception) exist within V-REP,
and have many different modes of operation (e.g.~through infra-red or sonar).
There is also a large degree of flexibility in specifying and controlling object
properties such as the object's center of mass, density, or mass itself.
Finally, materials in V-REP can also be customized, and properties such as the friction
value can be readily specified and changed on a whim. We outline assumptions we
made with regard to many of these properties in Table~\ref{tab:assumptions}.

\subsection{Grasp parameterization and gripper configuration}

We assume that all grasps can be parameterized in terms of a specific number of
contact points $\pmb{c} \in \mathbb{R}^3$ and contact normals $\pmb{n} \in
\mathbb{R}^3$. Let $\mathcal{G} = \{(\pmb{c},\pmb{n})~|~\pmb{c} \in
\mathbb{R}^3, \pmb{n} \in \mathbb{R}^3\}$ be the set of all grasps. Various
types of grasps can be simulated using different robotic hands. Both contact
positions and normals of the hand's fingertips are stored. The simulations
recorded in the dataset uses the Barrett Hand performing cylindrical precision
grasps, but note that the simulator can be used with any multi-fingered hand.

We model the hand as a free-floating entity unattached to any robotic arm, with
a proximity sensor attached to the hand's palm and aligned with the vector
normal (i.e.~pointing outwards). The proximity sensor serves two purposes: (1)
it sets a distance away from the object that the gripper is to be placed, and
(2) it permits verification that an object is in the line of sight of the hand.
We model the proximity sensor beam as a ray, but note that for interested users,
V-REP offers a variety of different modes including: pyramid, disc, cylindrical
and conical.

\subsection{Coordinate frames}

Let \{O\} be the object's body-attached coordinate frame, \{G\}  be the
body-attached coordinate frame on the manipulator's palm, \{W\} denote the
world coordinate frame, and \{T\} denote the body-attached  coordinate frame
of a table top located at ${^W}P_{T}$ = (\SIlist[list-units=single,list-final-separator =
{, }] {0.0; 0.0; 0.65}{\m})
\subsection{Object representation and properties}

We use the object dataset developed by Kleinhans \etal\cite{kleinhans2016G3DB},
that contains multiple object \textit{morphs} over a variety of object classes.
We use a subset of all available meshes, which were morphed with significant
differences between them. Each object has been pre-scaled and saved in the
Wavefront~.obj file format. Importing the file, we re-mesh, and assume that
within V-REP, the object is represented as a Complex shape. The Open Dynamics
Engine (ODE) is used for modeling gripper and object dynamics; while ODE was not
specifically designed for handling complex shapes, we found our simulation to be
fairly stable setting the number of allowable contacts to 64 and setting the
configuration to ``very accurate''.

The simulator allows assignment of a friction value for each object. A constant
friction value was assumed for each object. Furthermore, all objects were
assigned the same friction value. We also assumed that each object shared a
similar mass of \SI{1}{\kg}; while this assumption may not necessarily
correspond to real world phenomena (e.g.~where larger objects generally
correspond to greater mass), it is simple enough to change this to suit a
given purpose. Fixing the mass allowed us to make an assumption
of the grasp being employed; specifically, that a precision grasp can generate
enough force to equalize the object weight and lift the object. We assume
partial knowledge of the object's pose through a crude pose estimation technique
(Section~\ref{sec:object_pose}), which is employed for generating initial grasp
candidates.

\subsection{Vision sensors}

Two types of vision sensors are present in the scene: (RGB and Depth), as well
as a derived Binary mask for performing object segmentation\footnote{Note that all
vision sensors have an external dependency on OpenGL for rendering.}. For our
purpose, we assume that each camera can be physically placed coincident with
each other such that each of the collected images captures the same amount of
information, but through different modalities. In V-REP, we also ensure that
each camera only takes a single image (by setting the explicit handling property
to true) rather then streaming to avoid unnecessary computation.

Each camera is positioned a distance of \SI{0.25}{\metre} along the negative
Z-direction of the coordinate frame attached to the hand's palm, with
each camera sharing the same global orientation as the manipulator. In more technical
terms, this can be thought of as having a ``camera-in-hand'' configuration (such
as Baxter) and where the approach vector is along this line-of-sight. We use
perspective cameras, setting the resolution to be $128\times128$ (a modest size for
machine learning algorithms), the perspective angle of each camera to be
50$^{\circ}$, and near/far distance clipping planes of 0.01 and \SI{0.75}{\m}
respectively.

\begin{table}[htbp]
  \centering
  \caption{Overview of major parameters and assumptions}
    \begin{tabular}{ll}
    \toprule
    Component & Parameter \\
    \midrule
    Simulator & V-REP PRO EDU, Version 3.3.0 (rev. 0) \\
    Primary language & Lua \\
    Dynamics engine & ODE v0.12 \\
    Task  & Object pick from resting pose on table top \\
    Manipulator & Barrett Hand \\
    Vision types & RGB-D \\
    Object files & Kleinhans \etal~\cite{kleinhans2016G3DB} \\
    \toprule
    Component & Assumption \\
    \midrule
    Grasp type & Cylindric precision grasps \\
    Grasp candidates & Global and local rotations in object frame \\
    Grasp parameterization & Contact normals and positions in object frame \\
    Object mass & \SI{1}{\kg} \\
    Object pose & Coarse estimation (Section~\ref{sec:object_pose}) \\
    Object friction & 0.71 (default; constant among objects) \\
    Object geometry & Complex shape \\
    Vision perspective angle & $50^{\circ}$ \\
    Vision position & \SI{0.25}{\m} away from gripper position \\
    Vision orientation & Coincident with gripper orientation \\
    Vision clipping planes & Near: \SI{0.01}{\m}, Far: \SI{0.75}{\m} \\
    Vision resolution & $128\times128$ pixels \\
    \bottomrule
    \end{tabular}%
\label{tab:assumptions}%
\end{table}%

\section{Simulation initialization}
\label{sec:pipeline}

Each simulation requires an initial object and hand configuration: object
properties need to be defined, and a list of possible grasp candidates needs to
be generated.

\subsection{Initial Object and hand configurations and properties}
\label{sec:object_pose}

We begin by preprocessing all object meshes. Each object mesh is loaded into a
Python script, which makes use of the trimesh library \cite{trimesh} for
ensuring the meshes are watertight, and to obtain an estimate of the objects'
centers of mass and inertia.

Using these preprocessed values, we load each mesh into a V-REP simulation to
determine an initial resting pose for the object, and initial pose for a
gripper. We begin by assigning a bounding box for the object. This bounding box
is used to estimate the object's pose, by reorienting it with respect to
\{W\}, if not already aligned, and the frame center is assigned to be the
geometric center of the object. We then place the object \SI{0.3}{\m} along the
positive Z-direction of \{T\}, and is allowed to fall onto the table. Relative
to \{T\}, the object is then centered at $(x,y) = (0,0)$ using purely
translational components to maintain the resting pose.

Given this resting pose, we then place the gripper at an initial position along
the positive Z-direction in \{O\}. We chose this distance to be
$d=\sqrt{x^2+y^2+z^2}$~\SI{}{\m} away from the object's center, from the local
frame to the bounding box edges along the $x, y,$ \& $z$ directions
respectively. All object properties (including object pose, object bounding box,
and material) along with the gripper pose are recorded, and this process is
repeated for each object in the dataset.

\subsection{Grasp candidate database}
\label{sec:grasp_candidates}

In the grasping literature, a popular method of sampling grasp candidates is
through the use of surface normals emanating from the object
(e.g.~\cite{kappler2015leveraging},\cite{lenz2015deep}). This has been
implemented in simulators such as OpenRAVE~\cite{openRAVEGrasp}. The problem
with this approach is that there are several scenarios in which it may not
transfer well to the real world. Consider for example the following: (1)
sampling candidates from shiny or reflective surfaces (where it is difficult to
obtain object surface normals) and (2) sampling from areas with sharp edges and
acute angles between adjacent surfaces.

\begin{figure}
\centering
\begin{subfigure}{.24\textwidth}
  \centering
  \includegraphics[width=.85\linewidth]{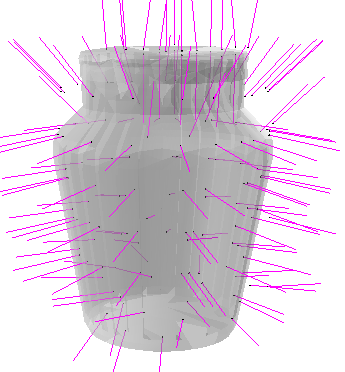}
  \caption{Grasp candidates generated \newline via surface normals}
\label{fig:hypothesis_normals}
\end{subfigure}%
\begin{subfigure}{.24\textwidth}
  \centering
  \includegraphics[width=.96\linewidth]{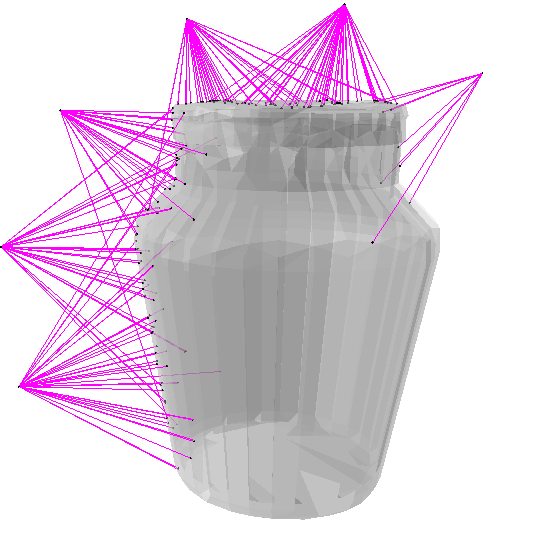}
  \caption{Grasp candidates generated via global and local rotations}
\label{fig:hypothesis_rotations}
\end{subfigure}
\caption{Different strategies for sampling grasp candidates.
\subref{fig:hypothesis_normals}) Grasp 
    candidates via surface normals; \subref{fig:hypothesis_rotations}) Grasp 
		candidates via global and local rotations of the gripper (with respect to
		the object). Purple lines denote the manipulator's approach vector. Only a
		subset of candidates are shown for clarity, and a 40\% transparency effect
		has been applied to the object.} 
\label{fig:hypothesis}

\end{figure}

The method used in this simulation to cover the possible grasp candidate space
is based on pre- and post-multiplication of the object configuration, which is
represented as a transformation matrix. Figure~\ref{fig:hypothesis} compares the
space covered by the proposed technique and a baseline which uses surface
normals. It can be seen in this figure that the method of pre- and
post-multiplication defines a sampling sphere around the center of the object.
While this resolves the above problems, using this method does require an
initial estimate of the object's pose. We estimate the object's pose according
to Section~\ref{sec:object_pose} above.

Given the object's bounding box and gripper pose, we calculate grasp candidates
offline by rotating the gripper globally (pre-multiply) and locally
(post-multiply) around the object. Following the convention in V-REP
\cite{vrepEulerAngles}, we multiply $3\times3$ rotation matrices in the order
$R_X(\alpha)R_Y(\beta)R_Z(\gamma)$, in the X, Y, and Z axes respectively.
Omitting $\alpha$, $\beta$, and $\gamma$ for clarity, the transformation matrix
is calculated according to:
\begin{equation}
    \label{eqn:multiply}
    Q = {R_X}{R_Y}{R_Z}~^{O}_{G}T~{R_X}{R_Y}{R_Z}
\end{equation}

\noindent where $Q$ represents the final transformation of the gripper
coordinate frame. Computing grasp-candidates is performed offline within a
Python script, and uses the estimated bounding box of the object, transformation
matrices ${}^{O}_{G}T$ and ${}^T_O{T}$\footnote{The complexity of a na\"{\i}ve
approach is $\mathcal{O}(n^6)$; but offline computation allows for grasp
candidates to be computed in parallel for each object being considered.}.
Formally, we exhaustively sample rotations following the constraints in Table
\ref{tab:rotations}. The constraints were chosen such that 8 rotations would
occur around the Z-axes (i.e.~every $45^{\circ}$), and local rotations would
occur on a slightly finer scale than the global rotations.

After computing Equation~\ref{eqn:multiply}, we check whether the new gripper
location is beneath the table or not (if so, we reject the grasp candidate), and
then solve a system of linear equations to check whether a vector normal from
the gripper's palm intersects with the object's bounding box. If this
intersection is true, we add the grasp candidate to the grasp-candidate database
and repeat the process until the list of rotations has been exhausted. Of all
the possible candidates in the database, we select up to $10,000$ to be verified
in the simulator.
%
\begin{table}[htbp]
  \centering
  \caption{Rotation constraints (in degrees)}
    \begin{tabular}{rccc}
    \toprule
    \multicolumn{1}{c}{Rotation} & Minimum & Maximum  & Increment \\
    \midrule
    \multicolumn{1}{l}{Global X} & 0     & 180   & 30 \\
    \multicolumn{1}{l}{Global Y} & 0     & 360   & 30 \\
    \multicolumn{1}{l}{Global Z} & 0     & 360   & 45 \\
    \multicolumn{1}{l}{Local X} & 0     & 180   & 20 \\
    \multicolumn{1}{l}{Local Y} & 0     & 360   & 20 \\
    \multicolumn{1}{l}{Local Z} & 0     & 360   & 45 \\
    \bottomrule
    \end{tabular}%
\label{tab:rotations}%
\end{table}%

\section{Simulation procedure}

\begin{figure}
	\centering
	\includegraphics[width=.95\linewidth]{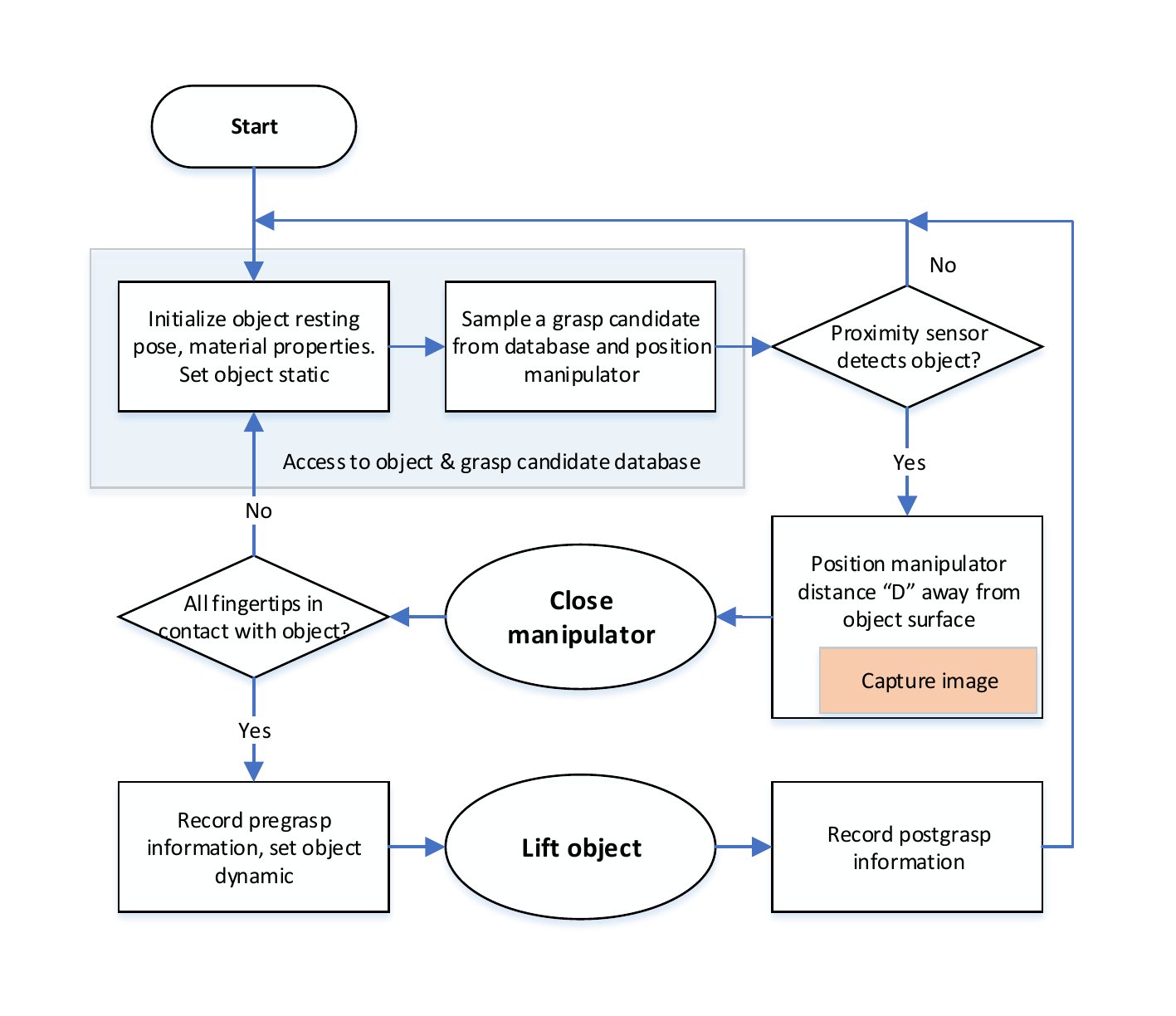}
	\caption{Flow-chart of grasp collection process}
\label{fig:collection_pipeline}
\end{figure}

The simulation procedure is illustrated in Figure~\ref{fig:collection_pipeline},
and begins by loading an object into the simulation, and initializing its mass,
inertia, and pose with values recorded during the initialization phase.  The
object is initially placed into a static state, such that when the fingertips
come into contact with the object, the object does not move.

After loading the object, the simulator samples a subset of the potential
candidates during the initialization phase (in this work, we use approximately
$1,500$ at a time) to test. A large majority of these grasps will be infeasible
due to gripper configurations and potential collisions with either the table or
object. In cases where this occurs, we stop the current attempt and move to the
next candidate.

\begin{figure*}
\centering
\begin{subfigure}{0.33\textwidth}
  \centering
	\includegraphics[width=0.99\linewidth]{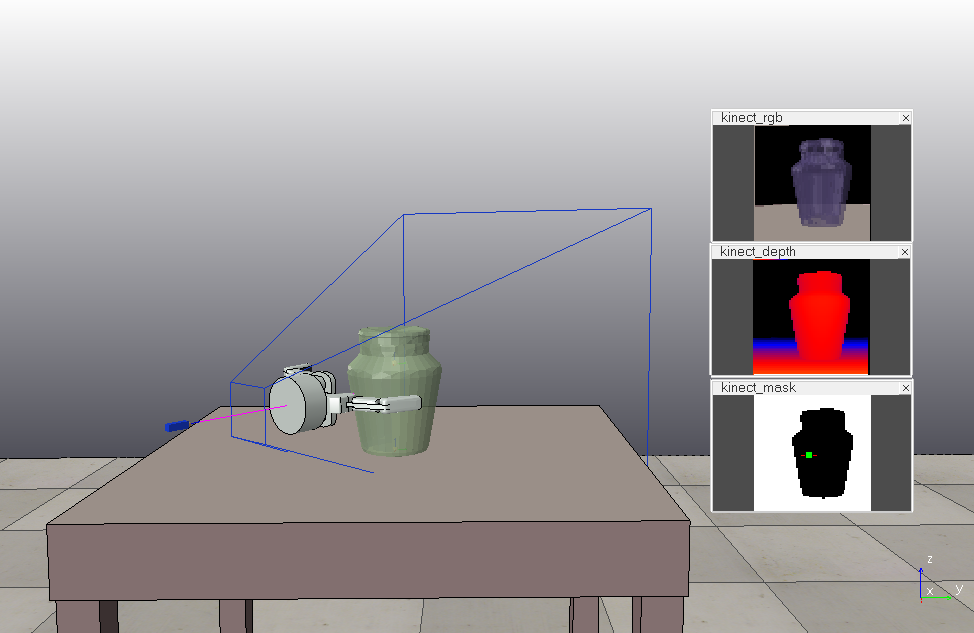}
    \caption{Positioning manipulator}
\label{fig:object_pregrasp}
\end{subfigure}%
\begin{subfigure}{0.33\textwidth}
  \centering
	\includegraphics[width=0.99\linewidth]{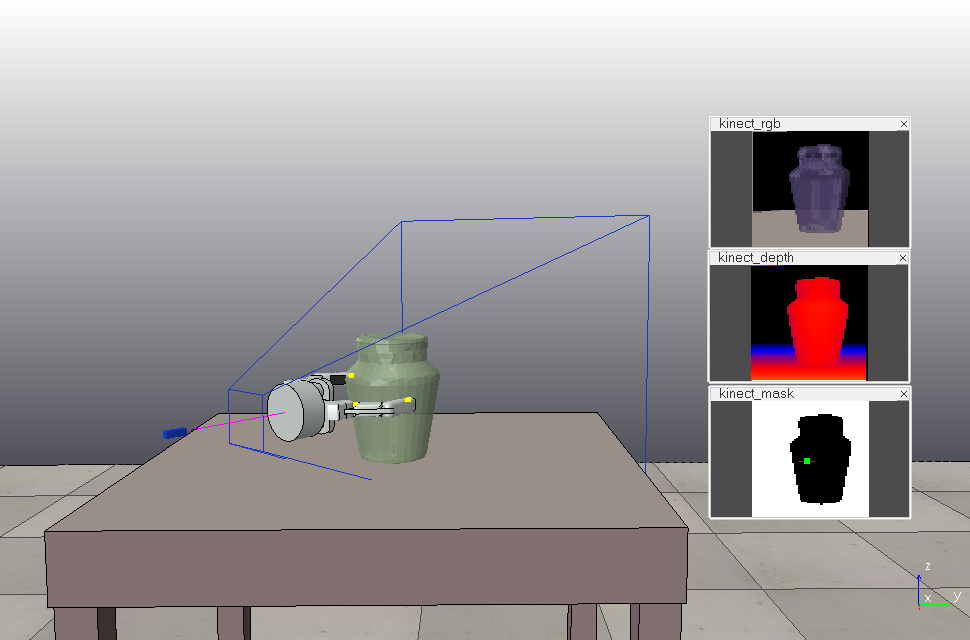}
    \caption{Grasping the object}
\label{fig:object_grasp}
\end{subfigure}%
\begin{subfigure}{0.33\textwidth}
  \centering
  \includegraphics[width=0.99\textwidth]{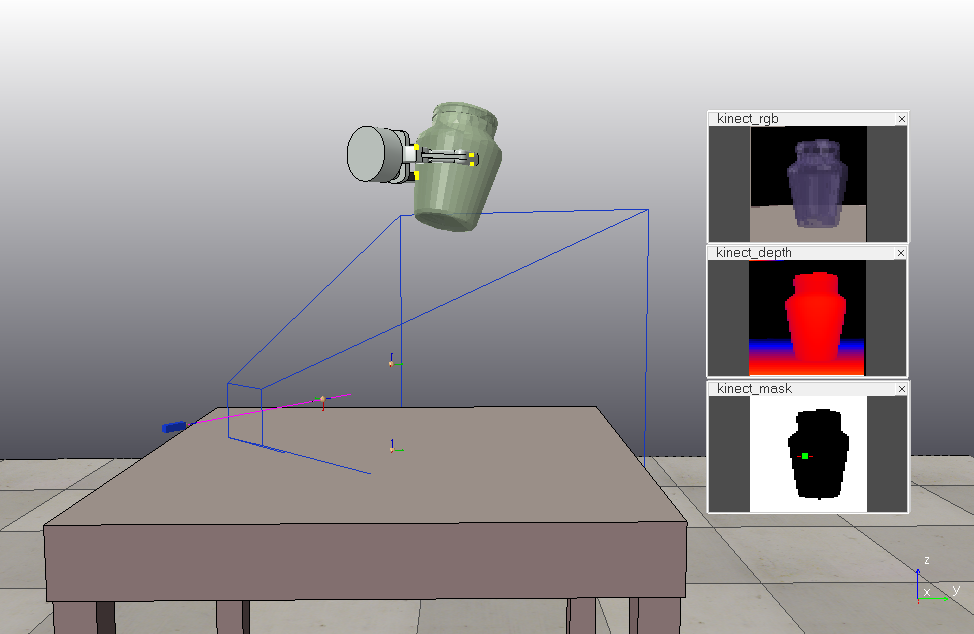}
  	\caption{Lifting the object}
\label{fig:object_lift}
\end{subfigure}
\caption{Sequence of actions for collecting grasps. Each image is depicted with
    the physial pose of the camera, with primary-viewing direction or
line-of-sight as a magenta line, along with the viewing angle in blue. As we
only take a single still image for each grasp attempt, the image as collected by
the cameras between different phases of the grasp are persistent.}
\end{figure*}

Each feasible grasp candidate is then checked using the proximity sensor in
order to verify the palm is facing the object. If, in this position, the
proximity sensor attached to the gripper detects an object, it records the
detected surface point and attempts three grasps (using the same gripper
orientation) at distances of: \SIlist[list-units=single,list-final-separator =
{, }] {0.06; 0.09; 0.12}{\m} away from the detected surface point and along the
original palm-normal (Figure~\ref{fig:object_pregrasp}). These distances were
chosen to lie within the distance between the gripper palm and fingertip
(\SI{0.145}{\m}) for the Barrett Hand, and allow us to explore the geometry of
the object at slightly different scales.
During each of the four attempts, the camera is positioned a distance of
\SI{0.25}{\m} away from the hand palm along the local negative Z-direction, and
records an image of the object before the grasp is
attempted\footnote{Computationally, the order the image is taken in is
irrelevant; in the real world, the image would be taken before the hand is
placed. In V-REP, we can explicitly set the focus of each camera to ignore
anything other than the object.}. Once the gripper has been placed and an image
recorded, the manipulator then closes around the object (Figure
\ref{fig:object_grasp}). If all fingertips are in contact with the object, the
object becomes dynamically simulated and the lift procedure begins.

We choose a target lift position of
(\SIlist[list-units=single,list-final-separator = {, }] {0.0; 0.0; 0.60}{\m})
relative to \{T\} and force the manipulator to maintain the current grasp pose
during travel. Once the gripper has reached the target location, if all
fingertips are still in contact with the object, the grasp is deemed
\textit{stable} and a success (Figure~\ref{fig:object_lift}). This procedure is
repeated until the list of grasp candidates has been exhausted. In V-REP, we
make use of the Reflexxes Motion Library~\cite{kroger2011opening} wrappers
(``simRMLxxx'' family) for computing the trajectory and for performing
incremental steps along the generated path.

%
\subsection{Different image and grasp mappings}

As the gripper was programmed to always close around the object in a similar
way, we found it interesting to collect two different views of the object during
the grasping process:

\begin{itemize}
	\item{Where the orientation of the camera always points upwards (one-to-many
	mapping), and}
	\item{Where the orientation of the camera always matches the orientation of the
	gripper (one-to-one mapping)}
\end{itemize}

The first point introduces ambiguity into the grasp space, by evoking a
one-to-many mapping between images and grasps. In this case, the gripper
orientation is not directly linked to the camera orientation, which means that a
single image may correspond to possibly many different grasps. The second point,
however, introduces a more direct relationship between images and grasps;
similar orientations of the object captured in the image reflect similar
orientations within the grasp. We have split this phenomenon into two separate
files for convenience.

\subsection{Parallelization}

Because such a large number of grasp candidates are sampled, and the number of
objects to be evaluated is relatively high, in order to create the dataset
within a feasible amount of time some form of parallelization is required.

The University of Guelph has a compute cluster consisting of 10 nodes, with each
node containing multiple Nvidia TITAN X GPUs and 2 Intel Xeon E5--2620 CPUs
running at 2.10GHz. Each CPU has 6 cores, and with hyperthreading gives us
access to 24 virtual cores and 64GB RAM. Using the grasp candidates sampled 
offline, as described in Section~\ref{sec:grasp_candidates}, we evenly
distribute the load across 4 compute nodes with 80 simulations running in
parallel. We use GNU Parallel for managing the load on each node
\cite{Tange2011a}.

We operate each scene in headless mode (i.e.~running without any graphical
interface), under an Xorg server due to requirements from the vision sensors
which require a small amount of memory from the graphics cards. In our
simulation, each scene typically uses around 4MiB, and for the Xorg server
around 20MiB. In total, we use slightly more than 100MiB of the graphics card
for running 20 simulations concurrently and have found this
process to take between 10--14 days to fully complete.

\section{Postprocessing}

Once all simulations have finished running, we apply a postprocessing step to
clean and standardize the collected data. This step consists of three parts: i)
decoding collected depth images, ii) automatic removal of grasp outliers, and
iii) manual inspection and final removal, which are performed before constructing
the dataset.

\subsection{Decoding depth images}

Within the simulation, information captured via a depth buffer is encoded to a
range between $[0, 1]$, and can be decoded to real-world values by applying:
\begin{equation}
\label{eqn:decode}
I = X_{\text{near}} + I*(X_{\text{far}}-X_{\text{near}})
\end{equation}

\noindent where $I$ is the collected image, and $X_{\text{near}}$,
$X_{\text{far}}$ are the near and far clipping planes respectively. Because some
of the images can be quite large, and depending on the view of the objects that
the cameras have, they may yield no useful shape information. In these
instances, the object typically occupies the full sensor resolution, and no
edges are visible. To combat this, we remove all object-grasp instance pairs
where the image variance is less then $1e^{-3}$. We also remove any grasps
where the collected image appears to bisect the table, which occurs when the camera
height matches that of the table height. Depth information encoded in this scenario 
is often at a minimum. 

\subsection{Postprocessing outliers}

When removing grasp outliers, we consider objects individually, and remove any
object-grasp instance pairs where one of the variables (either a fingertip
position or a normal) falls outside of 4 standard deviations of the population
mean. While on the surface a very simple method, we have found it to perform 
quite well in removing some of the more unlikely grasps, and reducing the 
number of grasps that will receive manual attention in the following step.

\subsection{Manual removal of physically inaccurate grasps}

We have noticed that grasps that make it through automatic filtering can often
be related to the complexity of the associated object mesh. Part of this is
linked to an earlier assumption that was made: specifically, that all meshes can
be simulated accurately within the environment as complex objects. However, all
objects are not created equal.

Objects with a greater number of faces, or those composed of several different
components have an (understandably) more difficult time within the simulator. In
addition to these difficulties, there are subtle cases where one of the
fingertips contacts an edge of the object. When this contact occurs, it is possible that
an improper contact normal may be retrieved during a dynamics pass within the
simulation. In all cases, we remove those grasps which we perceive as physically
impossible.

Once this is done, we consider all the objects for a given class, and remove one
to place into a test. From the remaining objects, we randomly sample 10\% of all
grasps to place into a validation set, while the remaining grasps comprise the training
set. We only populate the training, testing, and validation sets with
\textit{successful grasps}.

\begin{figure*}[htb!]
\centering
\colorbox{gray!10}{

\begin{subfigure}{0.33\textwidth}
  \centering
  {\includegraphics[width=0.9\linewidth]{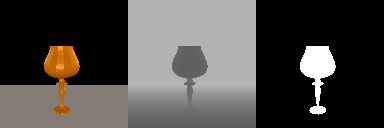}} 
  \caption{Object class 42: Wineglass}
\label{fig:class_wineglass}
\end{subfigure}%
\begin{subfigure}{0.33\textwidth}
  \centering
  {\includegraphics[width=0.9\linewidth]{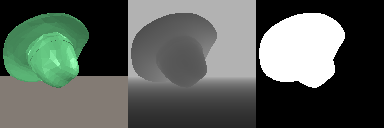}} 
  \caption{Object class 72: Hat}
\label{fig:class_hat}
\end{subfigure}
\begin{subfigure}{0.33\textwidth}
  \centering
  {\includegraphics[width=0.9\linewidth]{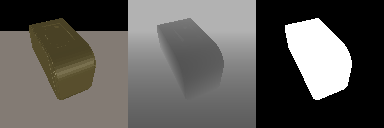}} 
  \caption{Object class 104: Toaster}
\label{fig:class_toaster}
\end{subfigure}
}

\colorbox{gray!10}{
\begin{subfigure}{0.33\textwidth}
  \centering
  \includegraphics[width=0.9\linewidth]{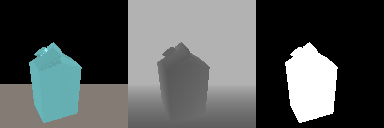} 
  \caption{Object class 50: Carton}
\label{fig:class_carton}
\end{subfigure}%
\begin{subfigure}{0.33\textwidth}
  \centering
  \includegraphics[width=0.9\linewidth]{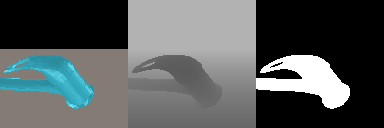} 
  \caption{Object class 53: Watertap}
\label{fig:class_watertap}
\end{subfigure}
\begin{subfigure}{0.33\textwidth}
  \centering
  \includegraphics[width=0.9\linewidth]{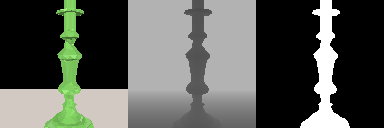} 
  \caption{Object class 54: Candlestick}
\label{fig:class_candlestick}
\end{subfigure}
}

\colorbox{gray!10}{
\begin{subfigure}{0.33\textwidth}
  \centering
  \includegraphics[width=0.9\linewidth]{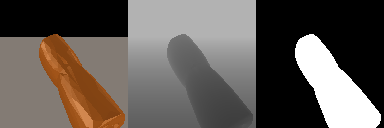} 
  \caption{Object class 59: Towel}
\label{fig:class_towel}
\end{subfigure}%
\begin{subfigure}{0.33\textwidth}
  \centering
  \includegraphics[width=0.9\linewidth]{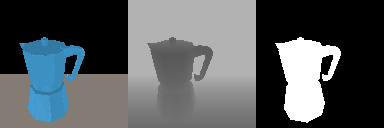} 
  \caption{Object class 65: Coffeemaker}
\label{fig:class_coffeemaker}
\end{subfigure}
\begin{subfigure}{0.33\textwidth}
  \centering
  \includegraphics[width=0.9\linewidth]{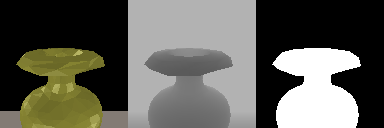} 
  \caption{Object class 71: Vase2}
\label{fig:class_vase2}
\end{subfigure}
}
  \caption{Sample images for different classes collected during simulation. Left: RGB, Center: depth image, Right: binary mask}
\label{fig:decoded_images}
\end{figure*}

\section{Dataset}

The code for this project can be accessed at
\hyperlink{https://github.com/mveres01/grasping}{https://github.com/mveres01/grasping},
while a sample dataset can be accessed at
\hyperlink{http://dx.doi.org/10.5683/SP/KL5P5S
}{http://dx.doi.org/10.5683/SP/KL5P5S}. The dataset has been saved in the HDF5
format, and was created using Python. While the postprocessing step removes most
abnormal grasps, note that there may be a select few that remain physically
inaccurate. Therefore the dataset is not without some noise. The statistics for
each of the training, testing, and validation sets are reported in
Table~\ref{tab:dataset_statistics}.
\begin{table}[htbp]
  \centering
  \caption{Dataset statistics. Note that the
	number of grasps may change upon final release of the dataset.}
    \begin{tabular}{lll}
    \toprule
    Element & \# Samples & \# Object classes \\
    \midrule
    Train & 32,100 & 20  \\
    Validation & 3,564 & 20 \\
    Test & 14,693 & 62 \\
    \bottomrule
    \end{tabular}%
\label{tab:dataset_statistics}%
\end{table}%
%
\subsection{Dataset overview}

Within each data split, there are three serialized data structures: images,
grasps, and object properties that help describe the current state of the
simulation and grasp process.

\verb images :~a 4-d array of images, in the format: (samples, channels, rows,
cols), where channels is composed of RGB-D elements.

\verb grasps :~a 2-d matrix of grasps : (samples, grasp), encoded with respect
to the camera frame.  Each grasp is encoded as the 18-dimensional vector $[\vec{p_1},
\vec{p_2}, \vec{p_3}, \vec{n_1}, \vec{n_2}, \vec{n_3}]$, where $p_i$ is the (x,
y, z) position of finger $i$ and $n_i$ is the (x, y, z) vector normal of finger
$i$.

\verb object_props :~a group of components, describing different aspects of the
grasping process. We focus mainly on static properties, and constrain this
primarily to frames of reference, and specific object properties. These are
defined further in subsections \ref{sec:object_props1}
\&~\ref{sec:object_props2}.

\subsection{Description of frames}
\label{sec:object_props1}

All frames are encoded as a $1 \times 12$ homogeneous transformation matrix. We
leave it to the user to format these as proper homogeneous transform matrices by
reshaping each matrix to be of shape $3 \times 4$, then adding the row vector
$[0, 0, 0, 1]$. Table~\ref{tab:simulation_frames} outlines the frames of
reference saved during data collection; note that ``workspace'' corresponds to
the frame \{T\} above.
\begin{table}[htbp]
  \centering
  \caption{Different frames used in the simulation}
    \begin{tabular}{lp{5.4cm}}
    \toprule
    Frame & Description \\
    \midrule
    \verb frame_cam2img_otm & estimated image frame with respect to camera frame \\
    \verb frame_cam2work_otm & {workspace frame with respect to camera frame, \newline one-to-many mapping }\\
    \verb frame_cam2work_oto & {workspace frame with respect to camera frame, \newline one-to-one mapping} \\
    \verb frame_work2cam_otm & {camera frame with respect to workspace frame,\newline one-to-many mapping} \\
    \verb frame_work2cam_oto & {camera frame with respect to workspace frame, \newline  one-to-one mapping}\\
    \verb frame_world2obj & object's (physical) reference frame with respect to world frame \\
    \verb frame_world2work & workspace frame (i.e.~center of the table top) with respect to world frame \\
    \bottomrule
    \end{tabular}%
\label{tab:simulation_frames}%
\end{table}%
%
%
\subsection{Description of object properties}
\label{sec:object_props2}

Several object-specific properties were also captured. These are summarized in
Table~\ref{tab:object_properties}.
\begin{table}[htbp]
  \centering
  \caption{Object properties}
    \begin{tabular}{lp{5.7cm}}
    \toprule
    Property & Description \\
    \midrule
    \verb object_name & name of the object  \\
    \verb work2com & {location of the object's center of mass with respect to workspace frame}\\
    \verb work2inertia & {object's inertia with respect to workspace frame} \\
    \verb work2mass & {object's mass with respect to workspace frame} \\
    \bottomrule
    \end{tabular}%
\label{tab:object_properties}%
\end{table}%
The object's center of mass is a $1 \times 3$ vector, inertia is a $ 1 \times 9$
vector, and mass is a single scalar.

\section{Conclusion}

In this paper, we presented an integrated system for collecting cylindrical
precision robotic grasps using the Barrett Hand and V-REP simulator. We demonstrated
an approach for computing grasp candidates using local and global rotations
around an object-centric reference frame, and presented our experience managing
large-scale data collection over multiple compute nodes. It is our hope that
other individuals are able to use these ideas in their own implementations.

\section*{Acknowledgment}

The authors would like to thank Brian Tripp and Ashley Kleinhans from the
University of Waterloo for early access to the object models in this paper, as
well as for discussion on specific design choices in a simulation similar to
ours. The authors would also like to thank Marc Freese from Coppelia Robotics
for technical support surrounding operation of the V-REP simulator.

\bibliographystyle{IEEEtran}
\bibliography{IEEEabrv,bib}

\appendix

\subsection{A note on verifying grasps}

In this work, we used the 18-dimensional vector containing contact positions and
normals to represent a grasp. In order to test predicted grasps within a
simulator, there are two potential options: Applying forces directly to the
object, or finding an optimal wrist pose and solving a series of inverse
kinematic equations to find the finger joint angles.

\subsubsection{Applying forces directly}
V-REP has the capability for applying arbitrary forces to an
object (e.g.~via the \hyperlink{http://www.coppeliarobotics.com/helpFiles/en/apiFunctions.htm\#simAddForce}{simAddForce} function),
which allows the user to circumvent the use of a robotic hand. This is likely the most direct
method for implementing into the current simulation, and would require swapping the
hand module for a module that reads in a set of contact positions and normals, and applies
them accordingly.

\subsubsection{Solving inverse kinematics}

In order to take advantage of the inverse kinematics modules within V-REP for
positioning the fingertips, a little help is needed to find the initial pose of
the manipulators wrist. This can be done by solving a series of linear
equations, making use of the Sequential Least Squares Programming implementation
in SciPy~\cite{slsqpScipy}.

Formally, we solve for an initial wrist position by optimizing the rotational
and translation components of the matrix $^{O}_{G}T$, minimizing the following
objective function \cite{ikhand}:
\begin{align*}
\min_{\alpha,\beta,\gamma,T_x,T_y,T_z} \sum_{i=1}^{N} {(C_i-Y_i)}^2
\end{align*}
where $\alpha, \beta, \gamma$ are the x, y, and z rotational components, $T_x$,
$T_y$, $T_z$ are the x,y, and z are the translational components, $N$ is the
number of fingertips, $C_{i}$ are the fingertip positions with respect to
\{O\} (obtained by multiplying ${}^{O}_{G}T$ with forward kinematics to the
manipulator's fingertips), and $Y_{i}$ is the predicted fingertip positions with
respect to \{O\}.

\end{document}